%% file: prdc21.tex
\documentclass[conference]{IEEEtran}
\IEEEoverridecommandlockouts

\usepackage{cite}
\usepackage{amsmath,amssymb,amsfonts}
\usepackage{algorithmic}
\usepackage{graphicx}
\usepackage[export]{adjustbox}
\usepackage{textcomp}
\usepackage[svgnames,table]{xcolor}

\include{config}

\include{scenario_tikz}

\usepackage{background}
\SetBgContents{Preprint}
\SetBgColor{gray}
\SetBgOpacity{0.2}
\SetBgScale{14}
\SetBgAngle{45}
\usepackage{tcolorbox}
\usepackage{fancyvrb}

\begin{document}
\begin{figure*}[ht]  
\begin{minipage}[t]{1\linewidth}
{
\Large
PREPRINT.\\
\\
This is a preprint version of the publication.\\
\\
See the PRDC'21 proceedings for the final version.
\\
}
\begin{tcolorbox}[title=Cite as,colback=white,left=5pt,right=5pt]
Klikovits S., Arcaini P. (2021) Handling Noise in Search-Based Scenario Generation for Autonomous Driving Systems. In: 26th {IEEE} Pacific Rim International Symposium on Dependable Computing, {PRDC} 2021, Perth, Australia, December 1-4, 2021. IEEE
\end{tcolorbox}

\begin{tcolorbox}[title=Bibtex,colback=white,left=0pt,right=0pt]
\begin{verbatim}
@inproceedings{KlikovitsArcaini:PRDC2021,
  author    = {Stefan Klikovits and Paolo Arcaini},
  title     = {Handling Noise in Search-Based Scenario Generation
for Autonomous Driving Systems},
  booktitle = {26th Pacific Rim International Symposium 
on Dependable Computing (PRDC)},
  publisher = {{IEEE}},
  year      = {2021},
}
\end{verbatim}
\end{tcolorbox}
\end{minipage}
\end{figure*}


\title{Handling Noise in Search-Based Scenario Generation for Autonomous Driving Systems\\
\thanks{The authors are supported by ERATO HASUO Metamathematics for Systems Design Project (No. JPMJER1603), JST. Funding reference number: 10.13039/501100009024 ERATO. S. Klikovits is also supported by Grant-in-Aid for Research Activity Start-up 20K23334, JSPS.}}

\author{
\IEEEauthorblockN{Stefan Klikovits}
\IEEEauthorblockA{
\textit{National Institute of Informatics}, Tokyo, Japan \\
klikovits@nii.ac.jp}
\and
\IEEEauthorblockN{Paolo Arcaini}
\IEEEauthorblockA{
\textit{National Institute of Informatics}, Tokyo, Japan\\
arcaini@nii.ac.jp
}
}

\maketitle

\begin{abstract}
This paper presents the first evaluation of \acl{knn}-Averaging (\knna) on a real-world case study.
\knna is a novel technique that tackles the challenges of noisy \ac{moo}.
Existing studies suggest the use of repetition to overcome noise.
In contrast, \knna approximates these repetitions and exploits previous executions, thereby avoiding the cost of re-running.
We use \knna for the scenario generation of a real-world \ac{ads} and show that it is better than the noisy baseline.
Furthermore, we compare it to the repetition-method and outline indicators as to which approach to choose in which situations.
\end{abstract}


\glsresetall

\section{Introduction}
Noisy system exections---\eg due to use of volatile communication means, complex and distributed system architectures, and non-deterministic execution times---are a challenging, but common characteristic of many modern \acp{cps}~\cite{AfzalICST20}.
In a growing number of systems, a given randomness cannot be avoided at all, for instance, when a system's architecture is built on asynchronous message passing between multi-process / multi-node systems.
The popular \ac{ros} framework is a well-known representative of this type of system, causing challenges throughout all phases of a \ac{cps}'s lifecycle, including development, testing, and deployment.

\begin{figure}
\centering
\scalebox{0.75}{
\begin{tikzpicture}

\draw[roadmarking,opacity=.5] (0,0) -- ++(1,0) ++(1,0) -- ++(4,0);
\draw[centreline,opacity=.5] (0,.5) -- ++(1,0) ++(1,0) -- ++(4,0);
\draw[roadmarking,opacity=.5] (0,1) -- ++(1,0) ++(1,0) -- ++(4,0);

\draw[roadmarking,opacity=.5] (1.0,0) -- ++(0,-3);
\draw[centreline,opacity=.5] (1.5,0) -- ++(0,-3);
\draw[roadmarking,opacity=.5] (2.0,0) -- ++(0,-3);

\draw[roadmarking,opacity=.5] (1.0,1) -- ++(0,1);
\draw[centreline,opacity=.5] (1.5,1) -- ++(0,1);
\draw[roadmarking,opacity=.5] (2.0,1) -- ++(0,1);

\node[rectangle,fill=black,minimum width=0.5cm,minimum height=3pt,anchor=north west,inner sep=0pt,opacity=0.5] at (1.5,0) {};
\node[stop,scale=1.5,opacity=0.5] at (0.75,-0.25) {};

\node[rectangle,fill=black,minimum width=0.5cm,minimum height=3pt,anchor=south west,inner sep=0pt,opacity=0.5] at (1,1) {};
\node[stop,scale=1.5,opacity=0.5] at (2.25,-0.25) {};

\node[rectangle,fill=black,minimum height=0.5cm,minimum width=3pt,anchor=north west,inner sep=0pt,opacity=0.5] at (2,1) {};
\node[stop,scale=1.5,opacity=0.5] at (0.75,1.25) {};

\node[rectangle,fill=black,minimum height=0.5cm,minimum width=3pt,anchor=south east,inner sep=0pt,opacity=0.5] at (1,0) {};
\node[stop,scale=1.5,opacity=0.5] at (2.25,1.25) {};

\node[ego,scale=1.5] (ego) at (1.75,-2.5) {};
\node[right=0 of ego] {\scriptsize \ego};
\draw[egopath] (ego) -- ++(0,4.5);

\node[vehicle,scale=1.5,rotate=90] (vehicle) at (5.5,0.75) {};
\node[above right=0 of vehicle] {\scriptsize \dynamicvehicle};
\draw[vehiclepath] (vehicle) -- ++(-5.5,0);

\node[circle,color=white,inner sep=0pt,minimum size=0.4cm,fill=ao] (trigger) at ($(ego)+(0,1)$) {\textbf{!}};
\node[left=0 of trigger] {\scriptsize \texttt{Trigger}};
\draw[-latex,draw=ao,dashed] (trigger) -- (vehicle) node[color=ao,pos=0.1,right]{$\Interval$ \scriptsize 3 sec.};

\end{tikzpicture}
}
\caption{\ego activates the timed trigger tries to cross after the stop. \dynamicvehicle starts on its trajectory and ignores the stop sign. 
We aim to adapt \dynamicvehicle's position and velocity to produce a collision.}
\label{fig:runstopsign-schema}
\end{figure}
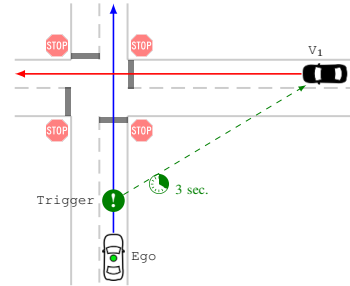

In our ongoing research on the testing of the Autonomoose\footnote{\url{https://www.autonomoose.net/}} \ac{ads}, we came across this problem when we applied \ac{sbt} using \ac{moo}~\cite{Goh2009} for the creation of crash scenarios.
Our goal here is to automatically adapt the parameters of an \ac{ads} scenario, \st the \ego vehicle collides with another road participant, \dynamicvehicle for instance (see \cref{fig:runstopsign-schema}). 
Subsequently, these crashes can be analysed by experts to identify whether \ego acted correctly or whether there is a problem.
Nonetheless, being based on \ac{ros}, our \ac{ads} simulator experiences noise and, therefore, suffers from the well-known problems of noisy search-based optimisation, namely the unwarranted preference of outliers and the selection of non-representative, volatile solutions~\cite{knnaveraging, goh2007investigation}.

\begin{figure}
\centering
\includegraphics[width=0.9\linewidth]{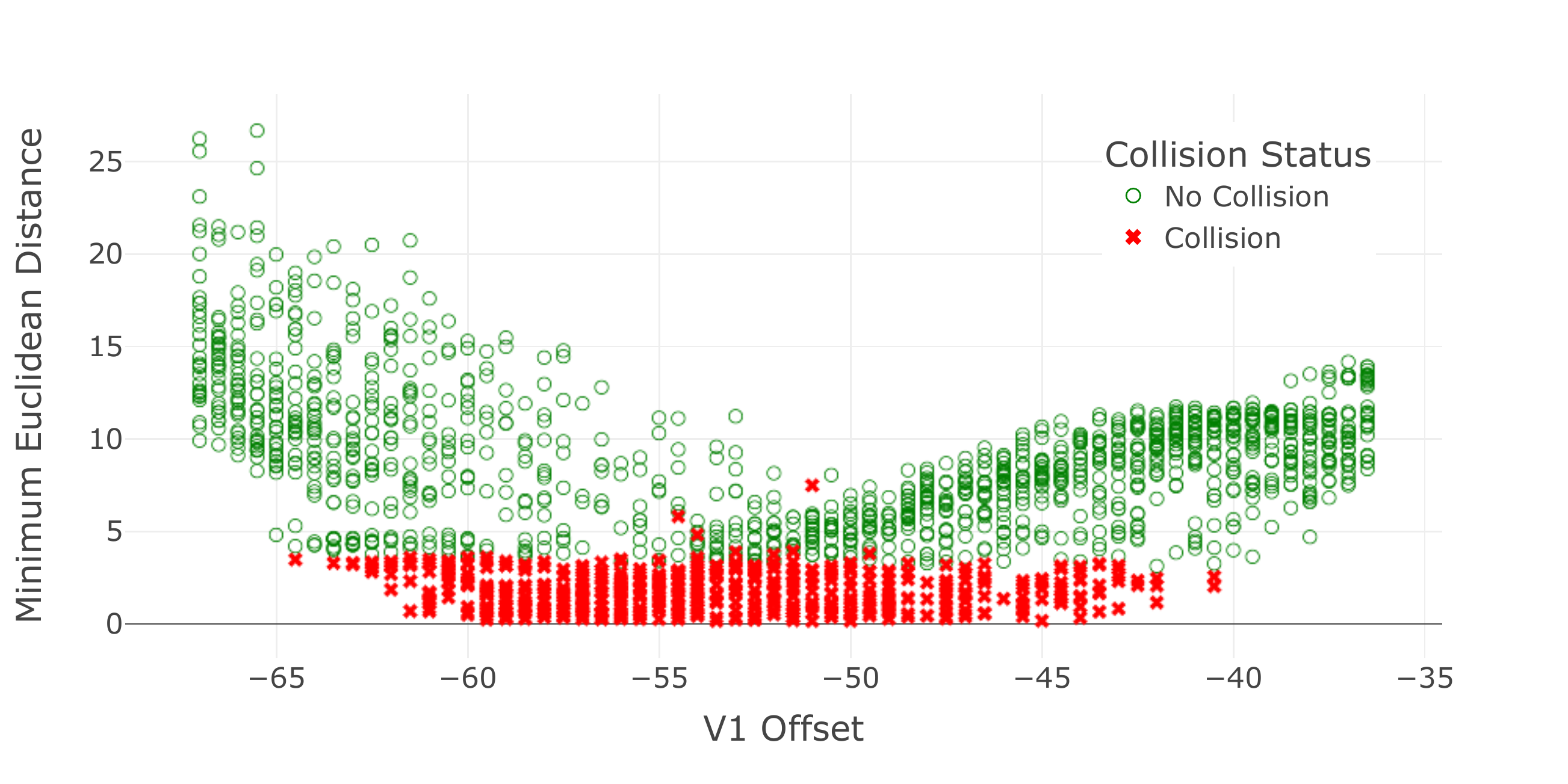}
\caption{Preliminary exploration showing noisy value spread and the inconsistent collisions in the search space.}
\label{fig:preliminary-distribution}
\end{figure}

The \ac{ads}'s noisy inter-process communication causes differing observations for repeated executions of the same setting, \st the distance between two cars may vary up to several meters.
To illustrate the effect and magnitude in our system, \cref{fig:preliminary-distribution} shows the results of an upfront evaluation of the scenario in \cref{fig:runstopsign-schema}. 
In this experiment, we systematically modify \dynamicvehicle's position in a value range where we suspect a collision. Each valuation is repeated 30 times where \cref{fig:preliminary-distribution} plots the valuation against the smallest Euclidean distance observed between \ego and \dynamicvehicle and marks whether a collision occurred.
As can be easily seen, not only do the ordinate-values differ vastly for a given situation, there is also not a single configuration that always leads to a collision. 
Logically, this problem becomes even more severe once the number of search variables is increased and the complexity raised to \ac{moo} for \ac{sbt}.

A naive solution to tackle the problem in \ac{moo} would be to repeat the evaluation of each configuration and calculate the average fitness value to account for the noise~\cite{Fitzpatrick1988}.
The problem, however, is that a typical simulation run takes roughly 1 to 2 minutes and requires a GPU. Assuming 1000 total optimisation steps (a typical \ac{sbt} setting), five-fold averaging would easily add some 80 to 160 hours of simulation time to each individual scenario search; 
or significantly lower the number of optimisation cycles, potentially causing worse results.
Both situations reduce this method's applicability for our purposes and broad application.

This paper describes the application of the \ac{knn}-Averaging~\cite{knnaveraging} technique to \ac{ads} scenario generation.
\knna is a recently proposed approach that avoids the costly need for re-running by relying on results from previous, neighbouring configurations. 
The advantage of \knna is that it increases the robustness of \ac{moo} searches, without requiring any additional simulation effort.
In~\cite{knnaveraging}, \knna showed good success when applied to synthetic benchmark problems from literature~\cite{ZDT2000}. 
In the rest of this paper, we show its viability on a real-world case study and compare it to the default, noisy setup and the naive repetition-solution.

\emph{Paper structure.}
\cref{sec:background} briefly discusses noisy \ac{moo} and re-states the \knna concept.
Next, \cref{sec:experimentDesign} introduces our experimental setup.
\cref{sec:expResults} outlines the results of the experiments and discusses the insights we draw.
\cref{sec:threatToValidity} explores threats to validity of our approach.
\cref{sec:related} explores related works, \cref{sec:conclusion} concludes and outlines future works.

\section{\acl{moo} \& \acl{knn}-Averaging} \label{sec:background}

\paragraph{\Acf{moo}~\cite{Goh2009}} \Ac{moo} refers to a collection of methods whose aim is to find a set of \emph{solution} vectors $\vecx \in \mathbb{X}$ which minimise the outcome of a set of \emph{objective} functions $f$.
The dimensions of $\mathbb{X}$ relate, for instance, to the position and velocity of \dynamicvehicle in \cref{fig:runstopsign-schema}.
\begin{equation}\label{eq:mooNoNoise}
\min_{\vecx \in \mathbb{X}} f(\vecx) = \{f_1(x), \dots, f_m(x)\}, \forall i \in \{1, \dots, m\}, f_i: \mathbb{X} \rightarrow \mathbb{R}
\end{equation}
The output $f(\vecx) \in \mathbb{Y}$ is called the \emph{objective} value of $\vecx$; $\mathbb{Y} \subseteq \mathbb{R}^m$ being the \emph{objective space}.
An optimisation problem is called \emph{multi-objective} if $m > 1$.

The complexity of the approach is that in many cases it is not possible to find a configuration that minimises all $f_i \in f$. As a result, we might observe a so-called \emph{Pareto front}, consisting of Pareto optimal points. 
These points are optimal in the sense that there is no point that minimises any of the objective dimensions without worsening another objective.
\cref{fig:pareto} displays the Pareto front of a search as red crosses \redcross.
Note, how for any \redcross there is no other \redcross whose $y_0$ value is lower, while its $y_1$ value is equally good or better, or vice-versa.
\cref{fig:pareto} also displays the ideal Pareto front \graypoint, representing the theoretically best values that can be obtained for the underlying ZDT1~\cite{ZDT2000} problem.
The goal of the \ac{moo} is to push the objectives \redcross as close as possible to the ideal Pareto front \graypoint.

\input{figures/pareto}

\paragraph{\Acp{ga}~\cite{Eiben2015}}
\Acp{ga} are a family of \acp{ea}. Their goal is to iteratively generate new \emph{populations} (sets of solutions) for a given number of \emph{generations}, by selecting good candidates and producing \emph{offspring} (combinations of two existing candidates) and \emph{mutants} (variants of one existing, good candidate). \Acp{ga} have been shown to be robust and continuously produce good results in various fields
such as evolutionary design, biological and chemical modelling, and artificial intelligence.
In this work, we apply the well-known NSGA-II algorithm~\cite{deb2002fast} implemented in the \texttt{pymoo} Python library~\cite{pymoo}.

\paragraph{Noisy \ac{moo}~\cite{Goh2009}}
Noisy \ac{moo} is a research domain that aims to overcome the problems introduced by non-deterministic objective functions. This means that for repeated executions, it is possible that some or all $f_i \in f$ may produce different results on exactly the same input \vecx.
In most studies, it is assumed that the noise is Gaussian-distributed~\cite{Goh2009}, which also seems to match the observations in the problems of the Autonomoose \ac{ads}.
When applying a \ac{ga} to noisy problems, we observe that it typically prefers outlier solutions, that have coincidentally good, non-representative objective values~~\cite{knnaveraging, goh2007investigation}.
When rerunning the yielded Pareto front's values, the \emph{effective} value (averaged over repeated runs) can be significantly different, and is typically worse than the one predicted by the \ac{ga}.
\cref{fig:pareto} shows an example of this behaviour, obtained from a noisy variant of the ZDT1 problem.
As can be seen, the noisy Pareto front \bluepoint is remarkably good. 
The effective Pareto front \bluestar for this experiment is significantly worse than the noisy one. Clearly, we see that the noisy Pareto front \bluepoint is non-representative.

\noindent\framebox{\parbox[t][2.0cm]{0.99\linewidth}{
The goal of our research is to reduce the effect of noise, such that the Pareto front yielded by the \ac{ga} is more representative, while still producing good effective results.
Casually speaking, we aim to push the \bluestar closer towards \graypoint, while at the same time reduce the length of arrows in \cref{fig:pareto}.
}}

\paragraph{\Ac{knn}-Averaging (\knna)~\cite{knnaveraging}}
\knna is a recently introduced method to overcome the problem of noisy \ac{moo}. 
Its goal is to approximate the repetition of $f(\vecx)$, without the actual need for rerunning expensive simulations.
Given an individual \vecx, \knna searches previously executed ``neighbour'' solutions (\ie close in solution space). Then, after obtaining the objective of \vecx (\eg through simulation), \knna calculates the mean of the new objective value with the neighbours' objectives and reports this as fitness to the \ac{ga}.

To avoid any bias on the search space dimensions and scaling, \knna uses \emph{standardised Euclidean distance} $\mathit{sed}(\vecx_1,\vecx_2)= \sqrt{\sum_{i=1}^N {\frac{(x_{1,i} - x_{2,i})^2}{\sigma_{i}^2}}}$, which relates all summands to their respective dimensional variance.

\knna can be configured using several hyperparameters. The first (obvious) one is the number of neighbours taken into account.
By convention, \knnAvg{1} represents the (noisy) baseline approach (called \baseline in the experiments), where only the noisy observation is used.
\knnAvg{10}, on the other hand, means that the algorithm will calculate the mean of the observation's value and those of the solution's nine closest neighbours.

Furthermore, a \emph{maximum-distance} can be configured to impose a limit on the distance of neighbours. This asserts that, especially in the beginning of the search or in generally sparsely populated areas of the search space, the algorithm does not use (probably) non-representative, distant solutions.


For more details, we refer to the original publication~\cite{knnaveraging}.

\section{Experiment design}\label{sec:experimentDesign}

\subsection{Experiment approaches}
We compare our \knna approaches \knnAvg{10} (with $k=10$ considered neighbours) and \knnAvg{50} ($k=50$) to the baseline approach \baseline that does not perform any mitigation action for noise, \ie the fully noisy approach. Moreover, we also compare them to the \emph{repetition approaches} \repAppr{2} and \repAppr{5}, which repeat each scenario configuration $n$ times ($n=2,5$) and use their mean objective values.

Similar to \knnAvg{k}, \baseline and \repAppr{} algorithms were implemented in \texttt{pymoo}, using the same \texttt{NSGA-II} optimisation algorithm, so rendering our results comparable.
As our goal is not to optimise the \ac{ga} parameterisation, but to study the \knna impact, we left the \ac{ga} in its default configuration. 
A study on any potential influence between \ac{ga} settings and \knna is left as future work.
For each experimental setting, the total budget \budget---\ie the total number of simulator executions---is equal, and so \baseline and both \knnAvg{}-settings cycled through twice as many optimisation generations as \repAppr{2}, and five times more generations than \repAppr{5}.
We believe that this configuration is practical and fair, since we assume a fixed total budget \budget as in realistic practical settings, and each approach decides how to allocate it.

\subsection{Definition of search objective and search spaces}
The specific search problem under study is the scenario shown in \cref{fig:runstopsign-schema}, for which we defined two search spaces.
Search space \searchSpace{1} is a simple version modifying \dynamicvehicle's \emph{position} and \emph{velocity} in the ranges $[-100,60]$ and $[35,65]$, respectively.
The position value indicates the offset from the scenarios's default position, corresponding to how many metres \dynamicvehicle is moved left or right in \cref{fig:runstopsign-schema}.
The absolute velocity is in km/h.

Search space \searchSpace{2} is more complex in several ways. 
First, it extends \searchSpace{1} and increases the \emph{velocity} range to $[3,100]$, and second,
it adds two more search variables, \emph{trigger delay} (in seconds; range $[0,10]$) and \emph{trigger position} (in metres; range $[-20, 20]$) which moves the trigger closer or further away from \ego's starting position.

Since our goal is to find avoidable collisions between the vehicles, we defined two objective functions that we compute from the simulation results.
\emph{Minimum Euclidean Distance} represents the closest distance of the two vehicles throughout the simulation. 
If this value is small enough, a crash is signalled by the simulator.
The second objective is to minimise the \emph{Velocity} of \dynamicvehicle as, clearly, collisions at fast speeds are more often unavoidable, while at low speeds \ego should be capable of making up for other driver's mistakes.

\subsection{Performance Indicators}\label{sec:performanceIndicators}
To quantify the benefit and allow objective comparisons between the \knna settings and the other compared approaches, we use the following performance indicators.

First, we use \meanErr as the \emph{mean distance} between a solution's calculated and effective objective values, as suggested in~\cite{fieldsend2005multi}.
Thus, \meanErr corresponds to the mean length of arrows in \cref{fig:pareto}.

Secondly, we calculate the hypervolume \HV~\cite{Li2019}, which calculates the volume between a given reference point in the objective space and the Pareto front's objective values.
By definition, the reference point is worse than all possible objective values and thus, the \ac{hv} increases as the solutions approach Pareto-optimality.
It is often used as a quality indicator when the ideal Pareto front is unknown.

To compare the difference of the \emph{predicted} (possibly noisy) and the \emph{effective} Pareto fronts, we will calculate two hypervolume values, \HVn and \HVeff. The \emph{predicted hypervolume} \HVn is based on the noisy predicted solutions, that may look exceptionally good only due to noise. Therefore, we also calculate the \emph{effective hypervolume} \HVeff, that is the hypervolume computed over the averaged values of 30 repetitions of each solution, that should give a more trustworthy indication of the quality of the final solutions. The difference between \HVn and \HVeff was already observed in~\cite{fieldsend2005multi}, although the authors used their proportion to judge the relative quality (similar to \meanErr).
As we also aim to compare against baseline approaches, we will use the actual \HV values.

\subsection{Experimental setup}
The experiments were conducted using the Autonomoose simulation platform, following the default installation and configuration\footnote{see \url{https://git.uwaterloo.ca/wise-lab/wise-sim-scenarios}}. No changes were made to the simulator or vehicle configuration, except the direct modification of the \texttt{nhtsa\_run\_stopsign} scenario during the search using prescribed solution values. 
The simulation runs were executed on AWS EC2 machines of flavour \texttt{g4dn.xlarge} under Ubuntu 18.04.

By following a guideline to conduct experiments with randomised algorithms~\cite{Arcuri2011}, each experiment was repeated ten times to account for the randomness of the search algorithm.\footnote{Due to cost and time constraints we could not repeat the experiment 30 times, but instead took particular care in comparing the results with appropriate statistical tests, as suggested in~\cite{Arcuri2011}.} We compare each pair of approaches in terms of \HVn, \HVeff, and \meanErr. Given a metric $M$ and two approaches $\mathrm{App}_1$ and $\mathrm{App}_2$, we compare the distributions of the values of $M$ using the Mann-Whitney U test\footnote{We use a non-parametric test because, in general, the compared values are not normally distributed.} for the statistical significance (at significance level $\alpha = 0.05$), and the Vargha-Delaney's \Atwelve as effect size. If the p-value of the Mann-Whitney U test is less than $\alpha$, we reject the null hypothesis and we claim that there is a significant difference; the \Atwelve value tells us which approach is higher and the strength of the difference.

To avoid interference between simulations, each simulation was run on its own machine on Amazon Web Services' EC2, avoiding interference of running parallel simulations, leading to a total of nearly 180 machine computation days for pure computation of the data.

\section{Experiment results}\label{sec:expResults}

\begin{figure*}[!tb]
\centering
\begin{subfigure}[b]{0.49\linewidth}
\centering
\includegraphics[width=0.70\linewidth,frame]{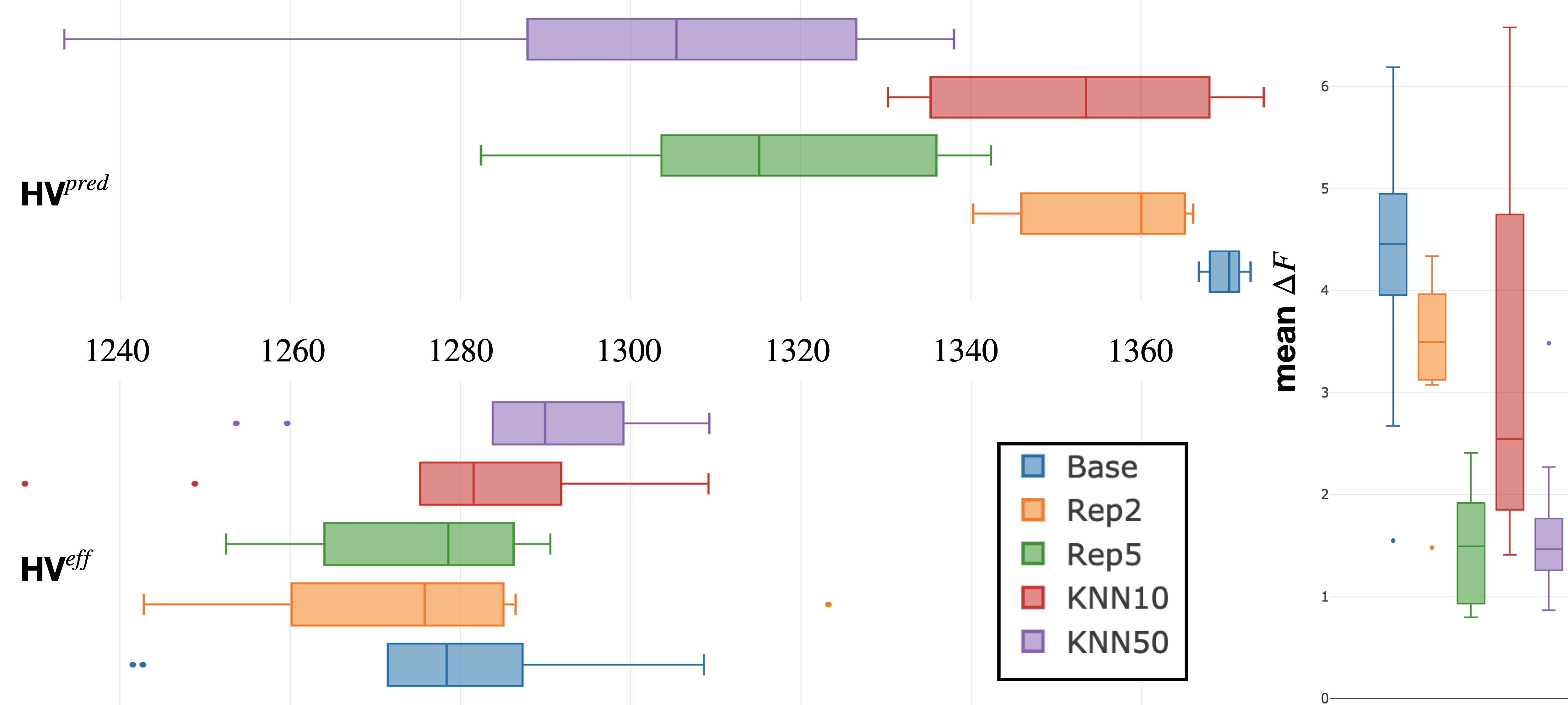}
\caption{\searchSpace{1} -- Population size: 10; Budget \budget: 500 simulations}
\label{fig:resultsPlotsSS1Pop10Gen50}
\end{subfigure}%
\begin{subfigure}[b]{0.49\linewidth}
\centering
\includegraphics[width=0.70\linewidth,frame]{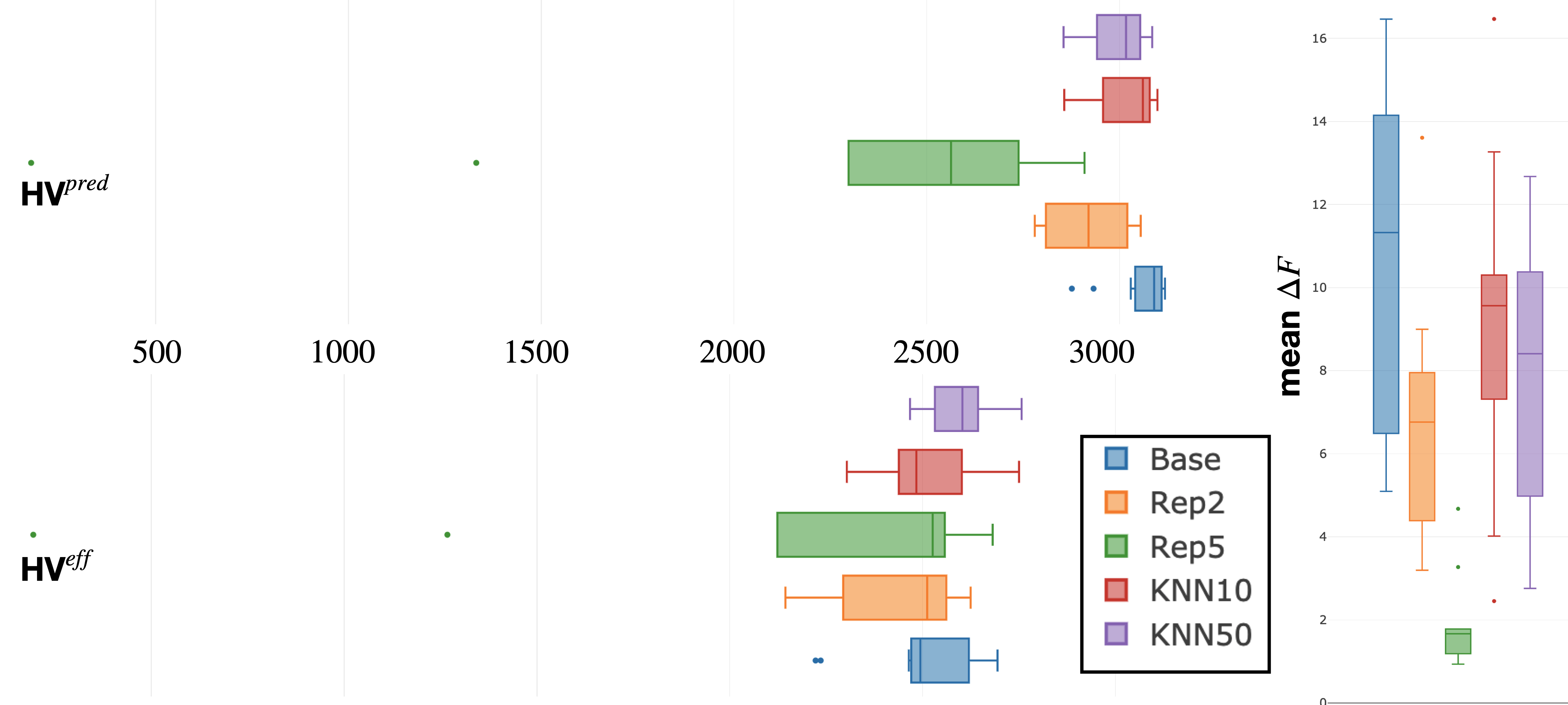}
\caption{\searchSpace{2} -- Population size: 10; Budget \budget: 500 simulations}
\label{fig:resultsPlotsSS2Pop10Gen50}
\end{subfigure}

\begin{subfigure}[b]{0.49\linewidth}
\centering
\includegraphics[width=0.70\linewidth,frame]{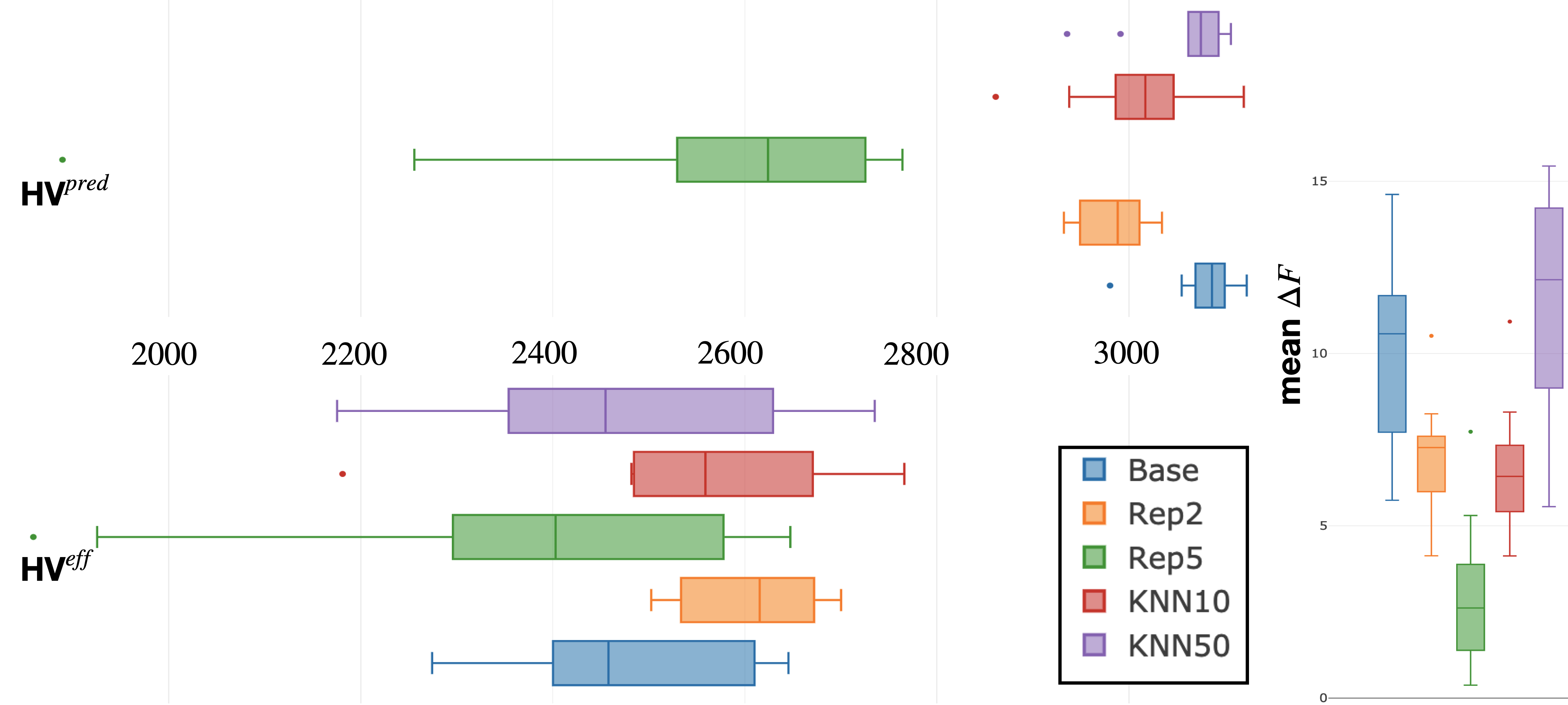}
\caption{\searchSpace{2} -- Population size: 20; Budget \budget: 500 simulations}
\label{fig:resultsPlotsSS2Pop20Gen25}
\end{subfigure}%
\begin{subfigure}[b]{0.49\linewidth}
\centering
\includegraphics[width=0.70\linewidth,frame]{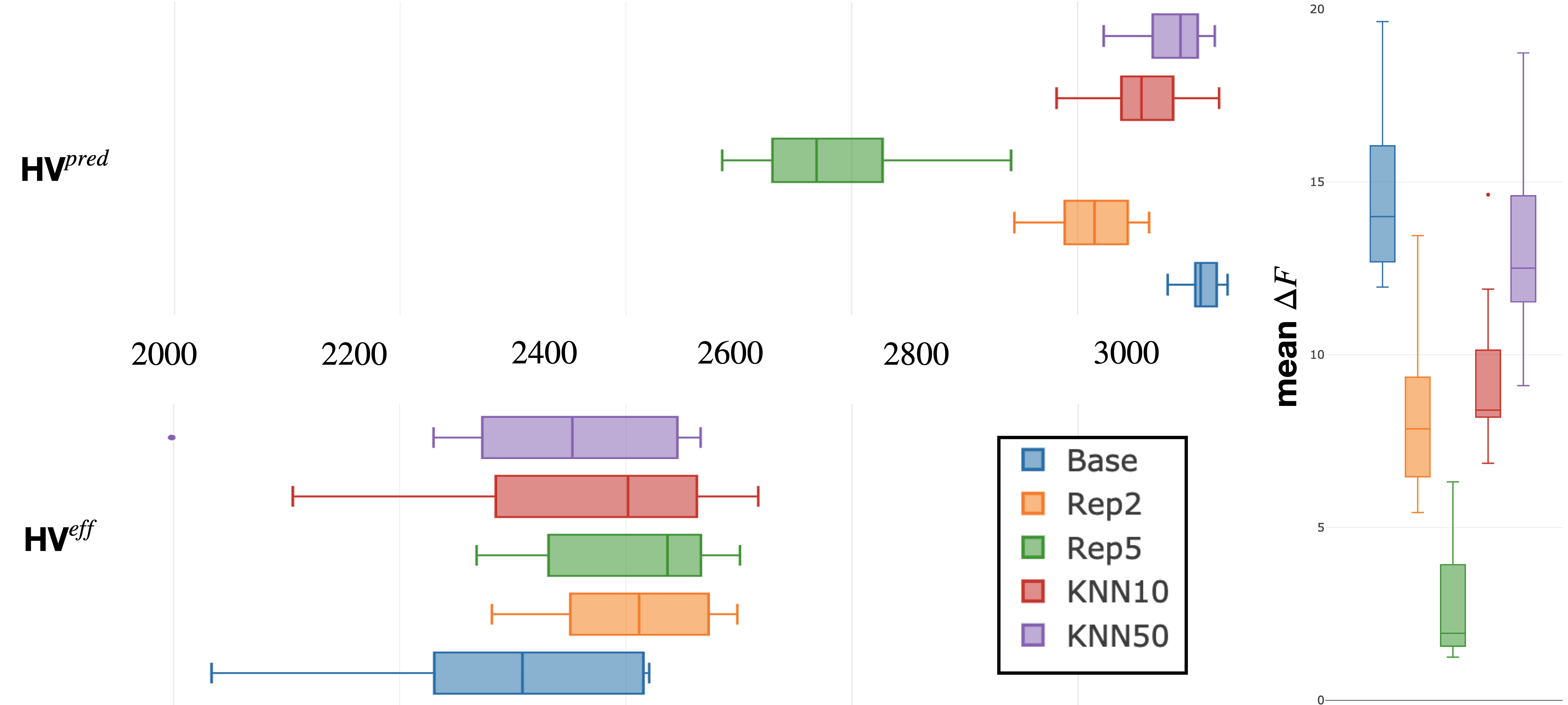}
\caption{\searchSpace{2} -- Population size: 20; Budget \budget: 1000 simulations}
\label{fig:resultsPlotsSS2Pop20Gen50}
\end{subfigure}
\caption{Boxplot representations of \HVn, \HVeff and \meanErr for the individual experiment settings. Note that the x-axes differ.}
\label{fig:resultsPlots}
\end{figure*}

We here introduce the experimental results, and discuss them using three research questions. Fig.~\ref{fig:resultsPlots} reports the distributions of the results of the three metrics for the five compared approaches.
\begin{table*}[!tb]
\centering
\caption{Comparison between the \ac{knn}-based settings \knnAvg{\selectedK} ($\selectedK = 10, 50$), the repetition approach \repAppr{n} ($n = 2, 5$), and the baseline approach \baseline (\better: $\mathrm{App}_1$ is statistically better. 
\worse: $\mathrm{App}_2$ is statistically better. 
\same: no significant difference.)}
\label{table:statResults}
\begin{subtable}[t]{0.24\textwidth}
\centering
\caption{\searchSpace{1} -- Pop 10; \budget 500}
\label{table:statResultsSS1Pop10Gen50}
\resizebox{\textwidth}{!}{
\begin{tabular}{llllll}
\toprule
$\mathrm{App}_1$ & $\mathrm{App}_2$ & \HVn & \HVeff & \meanErr\\
\midrule
\repAppr{2} & \baseline & \worse & \same & \same\\
\repAppr{5} & \baseline & \worse & \same & \better\\
\repAppr{5} & \repAppr{2} & \worse & \same & \better\\
\knnAvg{10} & \baseline & \worse & \same & \same\\
\knnAvg{10} & \repAppr{2} & \same & \same & \same\\
\knnAvg{10} & \repAppr{5} & \better & \same & \worse\\
\knnAvg{50} & \baseline & \worse & \same & \better\\
\knnAvg{50} & \repAppr{2} & \worse & \same & \better\\
\knnAvg{50} & \repAppr{5} & \same & \better & \same\\
\knnAvg{50} & \knnAvg{10} & \worse & \same & \better\\
\bottomrule
\end{tabular}
}
\end{subtable}
\begin{subtable}[t]{0.24\textwidth}
\centering
\caption{\searchSpace{2} -- Pop 10; \budget 500}
\label{table:statResultsSS2Pop10Gen50}
\resizebox{\textwidth}{!}{
\begin{tabular}{llllll}
\toprule
$\mathrm{App}_1$ & $\mathrm{App}_2$ & \HVn & \HVeff & \meanErr\\
\midrule
\repAppr{2} & \baseline & \worse & \same & \better\\
\repAppr{5} & \baseline & \worse & \same & \better\\
\repAppr{5} & \repAppr{2} & \worse & \same & \better\\
\knnAvg{10} & \baseline & \same & \same & \same\\
\knnAvg{10} & \repAppr{2} & \better & \same & \same\\
\knnAvg{10} & \repAppr{5} & \better & \same & \worse\\
\knnAvg{50} & \baseline & \worse & \same & \same\\
\knnAvg{50} & \repAppr{2} & \same & \better & \same\\
\knnAvg{50} & \repAppr{5} & \better & \better & \worse\\
\knnAvg{50} & \knnAvg{10} & \same & \same & \same\\
\bottomrule
\end{tabular}
}
\end{subtable}
\begin{subtable}[t]{0.24\textwidth}
\centering
\caption{\searchSpace{2} -- Pop 20; \budget 500}
\label{table:statResultsSS2Pop20Gen25}
\resizebox{\textwidth}{!}{
\begin{tabular}{llllll}
\toprule
$\mathrm{App}_1$ & $\mathrm{App}_2$ & \HVn & \HVeff & \meanErr\\
\midrule
\repAppr{2} & \baseline & \worse & \better & \better\\
\repAppr{5} & \baseline & \worse & \same & \better\\
\repAppr{5} & \repAppr{2} & \worse & \worse & \better\\
\knnAvg{10} & \baseline & \worse & \same & \better\\
\knnAvg{10} & \repAppr{2} & \same & \same & \same\\
\knnAvg{10} & \repAppr{5} & \better & \same & \worse\\
\knnAvg{50} & \baseline & \same & \same & \same\\
\knnAvg{50} & \repAppr{2} & \better & \same & \worse\\
\knnAvg{50} & \repAppr{5} & \better & \same & \worse\\
\knnAvg{50} & \knnAvg{10} & \same & \same & \worse\\
\bottomrule
\end{tabular}
}
\end{subtable}
\begin{subtable}[t]{0.24\textwidth}
\centering
\caption{\searchSpace{2} -- Pop 20; \budget 1000}
\label{table:statResultsSS2Pop20Gen50}
\resizebox{\textwidth}{!}{
\begin{tabular}{llllll}
\toprule
$\mathrm{App}_1$ & $\mathrm{App}_2$ & \HVn & \HVeff & \meanErr\\
\midrule
\repAppr{2} & \baseline & \worse & \same & \better\\
\repAppr{5} & \baseline & \worse & \better & \better\\
\repAppr{5} & \repAppr{2} & \worse & \same & \better\\
\knnAvg{10} & \baseline & \worse & \same & \better\\
\knnAvg{10} & \repAppr{2} & \better & \same & \same\\
\knnAvg{10} & \repAppr{5} & \better & \same & \worse\\
\knnAvg{50} & \baseline & \worse & \same & \same\\
\knnAvg{50} & \repAppr{2} & \better & \same & \worse\\
\knnAvg{50} & \repAppr{5} & \better & \same & \worse\\
\knnAvg{50} & \knnAvg{10} & \same & \same & \worse\\
\bottomrule
\end{tabular}
}
\end{subtable}
\end{table*}
Table~\ref{table:statResults} reports the comparison between all the approaches for the three metrics; \better means that the first approach is statistically better than the second approach, \worse that the second is better, and \same means that there is no difference.

\subsection{RQ1: Can \knna be better than the baseline approach and the repetition approach?}

In this RQ, we are interested in assessing whether the \knna is advantageous in test generation settings with limited budget. We have run, in the two search spaces \searchSpace{1} and \searchSpace{2}, all the approaches with population size 10, and a total budget \budget = 500 simulations. This means that \baseline, \knnAvg{10}, and \knnAvg{50} are run for 50 generations, \repAppr{2} is run for 25 generations, and \repAppr{5} is run for 10 generations.

Results are reported in Tables~\ref{table:statResultsSS1Pop10Gen50}-\ref{table:statResultsSS2Pop10Gen50} for the two search spaces. Regarding \HVn (\ie the hypervolume value computed on the final solutions), it is apparent that \baseline is better than any other approach. This is expected, as \baseline does not apply any technique to mitigate the noise (neither repeating nor k-nn averaging), and so noisy solutions that are exceptionally good will survive during the search and be returned as final solutions. We further notice that controlling the noise more (by more repetitions in the \emph{repetition approach} \repAppr{n}, or considering more neighbours in the proposed approach \knnAvg{k}) lead to worse \HVn results: \repAppr{5} is worse than \repAppr{2}, and \knnAvg{50} is worse than, or equivalent to, \knnAvg{10}. Again, this is reasonable, as the approaches that control the noise more are more precautios, and avoid to keep in the population very noisy solutions that are coincidentally very good. By comparing the repetition approach and our approach, it seems that there is no unique trend: \repAppr{2} can be worse than \knnAvg{10}, but can also be better than \knnAvg{50}; on the other hand, \knnAvg{10} and \knnAvg{50} are always better than, or equivalent to, \repAppr{5}.

We recall that \HVn is based on the computed solutions, that may look exceptionally good only due to noise; therefore, we also calculate the effective hypervolume \HVeff, that gives a real indication of the quality of the solutions (see \cref{sec:performanceIndicators}).
From Tables~\ref{table:statResultsSS1Pop10Gen50}-\ref{table:statResultsSS2Pop10Gen50}, we observe that, in this case, there is almost always no difference for \HVeff, even in the cases in which one approach was better for \HVn: this confirms that not controlling the noise can indeed produce noisy solutions that only look better, but are actually similar to other solutions when the noise is taken into consideration. A notable result is that the proposed approach \knnAvg{50} is actually always better than \repAppr{5}, and it was better or equivalent for \HVn.

As said before, the final solutions report fitness values that could be noisy, \ie they could deviate significantly from the real (effective) value. A good approach should return solutions whose computed fitness value is already close to the real value: such an approach is more trustworthy, as the user can have more confidence on the correctness of the results. Therefore, we here compare the different approaches w.r.t. the average error \meanErr of the returned solutions. From Tables~\ref{table:statResultsSS1Pop10Gen50}-\ref{table:statResultsSS2Pop10Gen50}, we observe that \repAppr{5} is better than \baseline and \repAppr{2}, showing that (as expected) simulating the solution multiple times leads to accurate results. We also observe that \knnAvg{10} provides worse results than \repAppr{5}, and is equivalent to \baseline and \repAppr{2}: this could mean that 10 neighbour solutions may not be enough to completely mitigate the noise. \knnAvg{50} is significantly better than \knnAvg{10}, \baseline, and \repAppr{2} in Table~\ref{table:statResultsSS1Pop10Gen50}, and equivalent to the same approaches in Table~\ref{table:statResultsSS2Pop10Gen50}. Note that \knnAvg{50} is once equivalent and once worse than \repAppr{5}; however, as we have observed before, it is always better than \repAppr{5} in terms of \HVeff: this could mean that \repAppr{5}, given this budget, is too precautios and is not able to find very good solutions, \ie it spends too much time in obtaining very accurate results, but it does not have time to explore more alternative solutions.

\subsection{RQ2: How do the results change by changing the configuration of the search algorithm?}

In a search algorithm, different hyperparameters (\eg population size, probability of crossover, etc.) can be selected, that affect the effectiveness of the approach. Since the ADS system under study is extremely costly, we can not perform a systematic study on the influence of the hyperparameters. Instead, we only perform a preliminary study, by increasing, in the test generation for search space \searchSpace{2}, the population size to $20$, by keeping the same budget \budget of 500 simulations. This means that \baseline, \knnAvg{10}, and \knnAvg{50} are run for 25 generations, \repAppr{2} is run for 13 generations, and \repAppr{5} is run for 5 generations. Experimental results are reported in Table~\ref{table:statResultsSS2Pop20Gen25}. By comparing the results with those of Table~\ref{table:statResultsSS2Pop10Gen50}, we observe that the performance of the proposed approaches \knnAvg{k} has in general decreased: \knnAvg{50} is no more better than \repAppr{2} and \repAppr{5} for \HVeff, and \knnAvg{50} is now also worse than \repAppr{2} for \meanErr. This can be explained by the fact that the accuracy of \knnAvg{k} improves with increasing generations (when solutions are converging towards optimum), while it is not particularly good for initial generations in which the solutions are more spread and so the averaging is less accurate.

\subsection{RQ3: How do the results change by increasing the budget?}

Another aspect that can be controlled in a search algorithm is the budget \budget. For ADS testing, having high budget is not feasible, due to the high cost of ADS simulations. In any case, we still try the experiments with a larger budget, by increasing it w.r.t. the one used in the experiment in RQ2. We run all the approaches using population size $20$, and $1000$ simulations as budget \budget. This means that \baseline, \knnAvg{10}, and \knnAvg{50} are run for 50 generations, \repAppr{2} is run for 25 generations, and \repAppr{5} is run for 10 generations. Experimental results are reported in Table~\ref{table:statResultsSS2Pop20Gen50}. As expected, \repAppr{n} still provides the best results in terms of \meanErr; however, we observe that \knnAvg{k} does not degrade its performance in terms of \HVeff, \ie there is no risk of obtaining more noisy results by running longer.

\subsection{Conclusions}
From the previous results, we observe that the proposed \knna approach is effective w.r.t. the baseline approach (no noise handling) and the \emph{repeat approach} (in which the noise is handle by multiple repetitions), when the budget is not too big (Figs.~\ref{fig:resultsPlotsSS1Pop10Gen50} and \ref{fig:resultsPlotsSS2Pop10Gen50}, and Tables~\ref{table:statResultsSS1Pop10Gen50} and \ref{table:statResultsSS2Pop10Gen50}), as it always the case in ADS testing that relies on costly simulations. Given the same budget, increasing the population size and reducing the number of generations does not seem to be effective (Fig.~\ref{fig:resultsPlotsSS2Pop20Gen25} and Table~\ref{table:statResultsSS2Pop20Gen25}). Of course, using a very large budget, the \knna and the \emph{repeat approaches} converge in terms of \HVeff (Fig.~\ref{fig:resultsPlotsSS2Pop20Gen50} and Table~\ref{table:statResultsSS2Pop20Gen50}); however, using such large budget is not feasible in the development practice, and so the repeat approaches are not a feasible option.

\section{Threats to Validity}\label{sec:threatToValidity}

The validity of our results could be affected by several threats. We discuss them in terms of \emph{construct}, \emph{conclusion}, \emph{internal}, and \emph{external validity}~\cite{Wohlin2012}.

{\it Construct validity.}
One such a threat is that the evaluation metrics are not suitable for evaluating the object of the investigation. In our context, we want to assess the ability of \knna to return good solutions in terms of the objective functions, but also that are trustworthy, \ie that are not simply good because they are noisy solutions. To assess this, the three metrics that we use in the experimental evaluation are suitable: \HVn measures the perceived quality of the (possibly noisy) solutions, \HVeff measures the real quality of the solutions after the noise has been removed, and \meanErr measures the distance between the returned objective value and the real one (\ie the one obtained with several repetitions).

{\it Conclusion validity.}
The ability to draw definitive conclusions could be affected by the random behaviour of search algorithms. Therefore, we repeated each experiment multiple times, as suggested by a guideline on conducting experiments with randomised algorithms~\cite{Arcuri2011}. Moreover, to draw significant conclusions, we have also compared the results of \knna and of the other compared approaches with suitable tests that account both for statistical significance and effect size.

{\it Internal validity.}
One such a threat is that the obtained results could be due to some other factor, rather than the applied technique. To mitigate such a threat, we thoroughly tested the implementation, and checked that bad solutions are not due to other factors such as the simulation timing out, or the simulator crashing.

{\it External validity.}
It could be that the effectiveness of the proposed approach does not extend to other benchmarks. First of all, the \knna approach was shown to be effective on several synthetic benchmarks in~\cite{knnaveraging}. In this paper, we show its applicability to a complex real world system. However, due to the cost of running the experiments, we could not assess on multiple search spaces with different objectives. Part of future work is to provide a more extensive evaluation.

\section{Related Work}\label{sec:related}

We here review works related to our approach, namely optimisation approaches handling noise, and search-based approaches for autonomous driving.

Some work have been proposed for handling noise in multi-objective optimisation; see~\cite{Goh2009} for a survey. For example, Fitzpatrick and Grefenstette use multiple evaluations of the fitness functions; we have used such approach as one of the compared approaches. Park and Ryu~\cite{ParkGECCO2011}, instead, handle the noise by doing multiple evaluations of the solutions over several different generations. Some works, instead, modify the ranking methods to take into account the noise, as done, for example, by Hughes~\cite{Teich2001} and Teich~\cite{Teich2001}.

Search-based testing has been extensively applied to ADSs. For example, Li et al.~\cite{LiISSRE20} propose AV-FUZZER, an SBT approach combined with fuzzing, that has the aim of finding safety violations of an ADS implemented in Apollo. Ben Abdessalem et al.~\cite{Abdessalem2016ASE} used multi-objective search combined with \emph{surrogate models} to test a Pedestrian Detection Vision based (PeVi) system. Gambi et al.~\cite{gambi2019automatically} search for road scenarios in which the lane keeping component of the ADS fails and the autonomous vehicle drives off the road. Cal\`{o} et al.~\cite{calo2020Generating}, instead, try to find collision scenarios that are \emph{avoidable}, \ie in which the ADS configured in a different way would not lead to collision.

\glsresetall

\section{Conclusions \& Future Work}\label{sec:conclusion}
This paper presents the results of applying \knna in noisy \ac{moo}.
Our experiments---stemming from a real-world case study on designing \acs{ads} scenarios---show that \knna clearly outperforms the noisy baseline setting.
Furthermore, we show that in certain settings \knna also outweighs the naive approach of repetition.
This shows that \knna is a valid alternative to the repetition approach, especially in situations with limited budget.

Nonetheless, we observe that the choice of search hyperparameters (\eg population size) has a big impact on the performance of \knna and the repetition.
Our next steps include a deeper study into various configurations such that we might provide best-practices for hyperparameter selection.
Next, we aim to extend our evaluation to additional case studies. 
As our time and budget constrain our work, we are additionally looking for other noisy real-world systems with less execution time.
Finally, we are in the process of integrating \knna into our general \ac{sbt} and scenario generation workflow. 
To this end, we are developing an extension of the \texttt{pymoo} library, that allows \knna's convenient off-the-shelve use.


\bibliographystyle{IEEEtran}
\bibliography{bibliography}

\end{document}

%% file: config.tex
\usepackage[svgnames,table]{xcolor}
\definecolor{darkblue}{rgb}{0, 0, 0.7}

\usepackage{verbatim}

\usepackage{amsfonts}
\usepackage{amssymb}
\usepackage{amsmath}
\usepackage{times}
\usepackage{stfloats}

\usepackage{pgfplots}  

\usepackage{caption}
\usepackage{subcaption}
\usepackage{pifont}
\usepackage{booktabs}

\usepackage{wrapfig}

\usepackage{paralist}

\usepackage{listings}
\usepackage[colorlinks=true,
pagebackref=false,
urlcolor=blue!50!blue,
citecolor=blue!50!blue,
linkcolor=darkblue]{hyperref}

\usepackage[capitalise,nameinlink]{cleveref}
\crefname{lstlisting}{listing}{listings}
\Crefname{lstlisting}{Listing}{Listings}
\crefname{line}{Line}{Lines}
\Crefname{line}{L}{L}
\crefname{section}{§}{§§}
\Crefname{section}{§}{§§}

\usepackage{xspace} 

\newcommand{\eg}{e.g.\xspace}

\newcommand{\ie}{i.e.\xspace}
\newcommand{\st}{s.t.\xspace}

\newcommand{\dynamicvehicle}{\texttt{V$_1$}\xspace}
\newcommand{\ego}{\texttt{Ego}\xspace}

\newcommand{\vecx}{\ensuremath{\vec{x}}\xspace}

\newcommand{\knna}{kNN-Avg\xspace}

\newcommand{\knnAvg}[1]{\ensuremath{\mathtt{KNN}_{#1}}\xspace}
\newcommand{\baseline}{\ensuremath{\mathtt{B}}\xspace}
\newcommand{\repAppr}[1]{\ensuremath{\mathtt{Rep}_{#1}}\xspace}

\newcommand{\HV}{HV\xspace}
\newcommand{\HVeff}{HV$^{\mathit{eff}}$\xspace}
\newcommand{\HVn}{HV$^{\mathit{pred}}$\xspace}

\newcommand{\budget}{\ensuremath{\mathit{Bgt}}\xspace}

\newcommand{\searchSpace}[1]{\ensuremath{\mathit{SS}_{#1}}\xspace}

\definecolor{badred}{RGB}{240,0, 0}
\definecolor{goodgreen}{RGB}{0, 216, 0}

\newcommand{\meanErr}{\ensuremath{\Delta F}\xspace}
\newcommand{\better}{\textcolor{goodgreen}{\ding{51}}\xspace}
\newcommand{\worse}{\textcolor{badred}{\ding{55}}\xspace}
\newcommand{\same}{\textcolor{orange}{\ensuremath{\equiv}}\xspace}
\newcommand{\Atwelve}{\^{A}\textsubscript{12}\xspace}

\newcommand{\selectedK}{\ensuremath{k}\xspace}

\newcommand{\graypoint}{\protect\tikz{\protect\node[circle,fill=gray,opacity=0.5,inner sep=0pt,minimum size=4pt] {};}\xspace}
\newcommand{\redcross}{\protect\tikz{\protect\node[color=red,opacity=0.75,mark size=3pt, line width=1pt,yshift=0.5em]{\protect\pgfuseplotmark{x}};}\xspace}
\newcommand{\bluepoint}{\protect\tikz{\protect\node[color=blue,opacity=0.35,mark size=1.5pt] {\protect\pgfuseplotmark{*}};}\xspace}
\newcommand{\bluestar}{\protect\tikz{\protect\node[color=blue,opacity=0.75,mark size=2pt,line width=1pt] {\protect\pgfuseplotmark{asterisk}};}\xspace}

\usepackage[acronyms,shortcuts,section=section,style=super,nopostdot,nogroupskip,nomain,nonumberlist]{glossaries}
\glsdisablehyper

\newacronym{ads}    {ADS}   {autonomous driving system}
\newacronym{cas}    {CAS}   {collision avoidance system}
\newacronym{cps}    {CPS}   {cyber-physical system}
\newacronym{ea}     {EA}    {evolutionary algorithm}
\newacronym{ed}     {ED}    {Euclidean distance}
\newacronym{ga}     {GA}    {genetic algorithm}
\newacronym{gd}     {GD}    {generational distance}
\newacronym{hv}     {HV}    {hypervolume}
\newacronym{igd}    {IGD}   {inverse generational distance}
\newacronym{iot}    {IoT}   {internet of things}
\newacronym{knn}    {kNN}   {k-nearest neighbours}
\newacronym{moo}    {MOO}   {multi-objective optimisation}
\newacronym{mop}    {MOP}   {multi-objective problem}
\newacronym{mosp}    {MOSP}   {multi-objective search problem}
\newacronym{qi}     {QI}    {quality indicator}
\newacronym{ros}    {ROS}   {Robot Operating System}
\newacronym{sbt}    {SBT}   {search-based testing}
\newacronym{sed}    {SED}   {standardised Euclidean distance}

%% file: scenario_tikz.tex
\usepackage{tikz}

\usepackage[clock]{ifsym} 

\usetikzlibrary{positioning}
\usetikzlibrary{calc} 
\usetikzlibrary{arrows}
\usetikzlibrary{arrows.meta}
\usetikzlibrary{shapes}
\usetikzlibrary{shapes.arrows}
\usetikzlibrary{patterns}

\definecolor{ao}{rgb}{0.0, 0.5, 0.0}
\pgfmathsetmacro{\markingwidth}{1pt}

\tikzset{
    roadmarking/.style={draw=gray!80,line width=\markingwidth},
    centreline/.style={roadmarking,dashed,dash pattern=on 7pt off 4pt},
    stopline/.style={roadmarking,line width=4pt},
    basevehicle/.style={rectangle,minimum height=0.5cm,minimum width=0.25cm,inner sep=0pt,},
    vehicle/.style={basevehicle,execute at begin node={\includegraphics[trim=0.5cm 0 0.5cm 0,clip,width=.25cm]{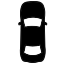}}},
    parkedvehicle/.style={basevehicle,execute at begin node={\includegraphics[trim=0.5cm 0 0.5cm 0,clip,width=.25cm]{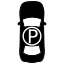}}},
    ego/.style={basevehicle,execute at begin node={\includegraphics[trim=0.5cm 0 0.5cm 0,clip,width=.25cm]{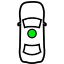}}},
    pedestrian/.style={minimum width=0.2cm, minimum height=0.2cm,inner sep=0cm,execute at begin node={\includegraphics[width=.3cm]{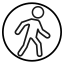}}},
    egopath/.style={draw=blue,-latex,thick},
    vehiclepath/.style={draw=red,-latex,thick},
    pedestrianpath/.style={draw=ao,-latex,thick},
    crosswalk/.style={rectangle,fill=gray!80,minimum height=0.06cm,minimum width=0.35cm,inner sep=0},
    stop/.style={minimum height=0.1cm,execute at begin node={\includegraphics[trim=0.5cm 0 0.5cm 0,clip,width=.25cm]{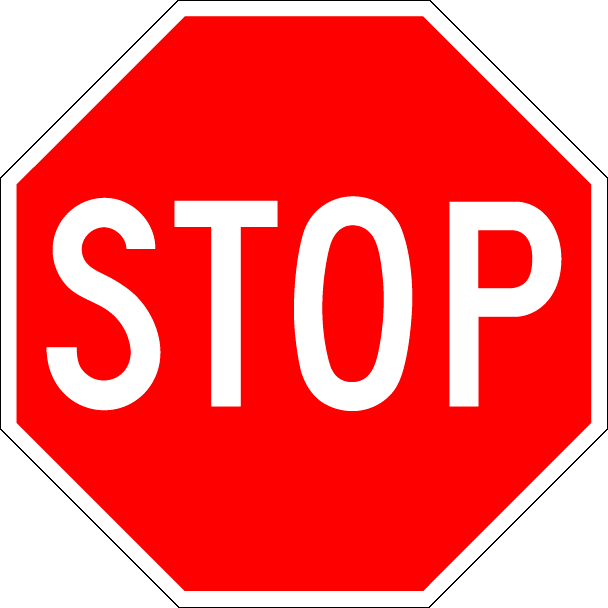}}},
    bump/.style={minimum height=0.1cm,execute at begin node={\includegraphics[trim=0.5cm 0 0.5cm 0,clip,width=.25cm]{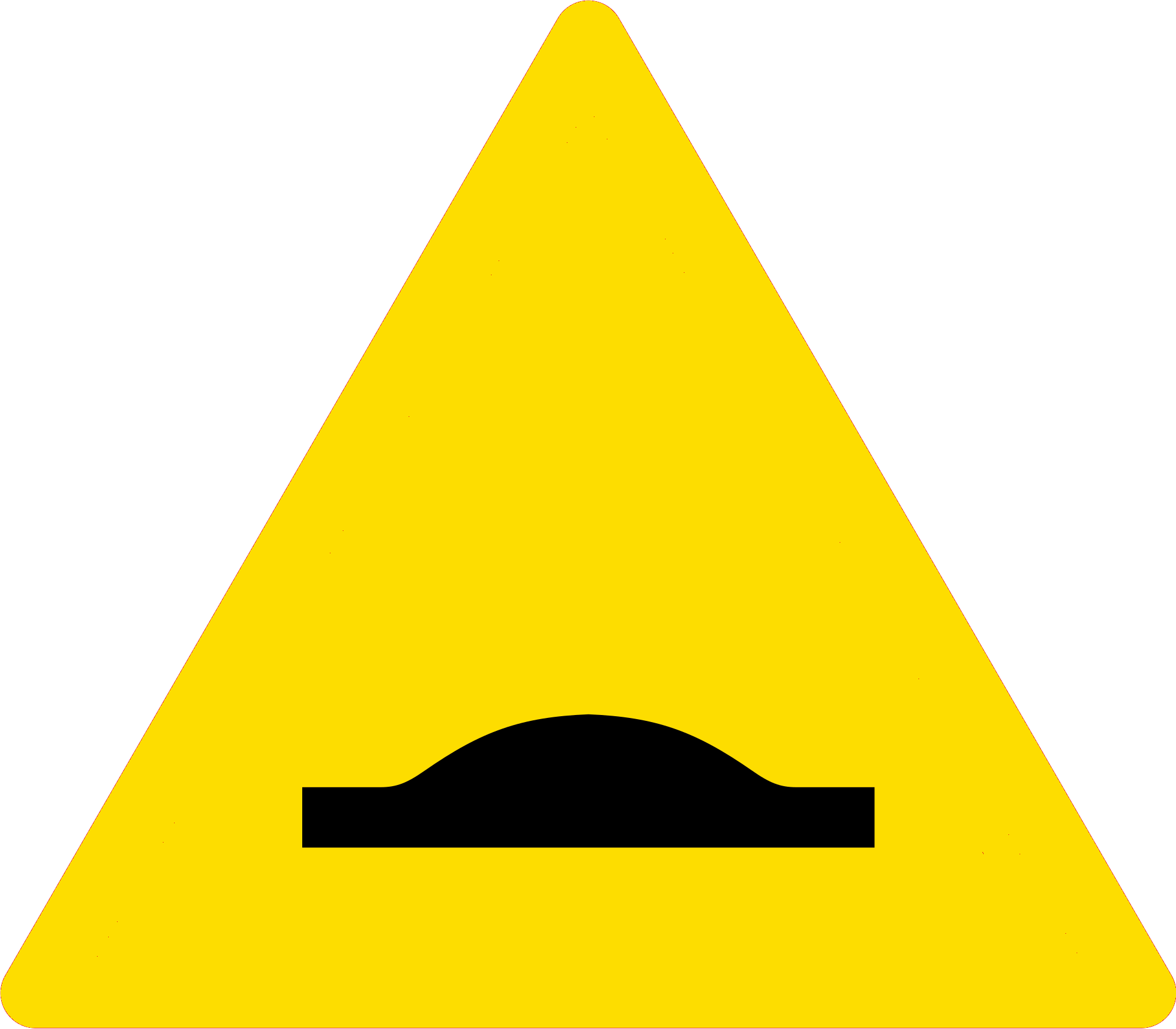}}},
    trigger/.style={circle,color=white,inner sep=0pt,minimum size=0.25cm,fill=ao,execute at begin node={\textbf{!}}}
}

%% file: figures/pareto.tex
\def\zdttruepareto{
(0.02020202,0.85786619), 
(0.04040404,0.79899244), 
(0.06060606,0.75381702), 
(0.08080808,0.71573238), 
(0.1010101 ,0.68217914), 
(0.12121212,0.65184469), 
(0.14141414,0.62394928), 
(0.16161616,0.59798487), 
(0.18181818,0.57359857), 
(0.2020202 ,0.55053343), 
(0.22222222,0.52859548), 
(0.24242424,0.50763404), 
(0.26262626,0.48752926), 
(0.28282828,0.46818398), 
(0.3030303 ,0.44951812), 
(0.32323232,0.43146476), 
(0.34343434,0.41396728), 
(0.36363636,0.39697731), 
(0.38383838,0.38045308), 
(0.4040404 ,0.36435827), 
(0.42424242,0.34866105), 
(0.44444444,0.33333333), 
(0.46464646,0.31835019), 
(0.48484848,0.30368938), 
(0.50505051,0.28933095), 
(0.52525253,0.27525692), 
(0.54545455,0.26145105), 
(0.56565657,0.24789857), 
(0.58585859,0.234586  ), 
(0.60606061,0.22150106), 
(0.62626263,0.20863243), 
(0.64646465,0.19596975), 
(0.66666667,0.18350342), 
(0.68686869,0.17122459), 
(0.70707071,0.15912503), 
(0.72727273,0.14719713), 
(0.74747475,0.13543378), 
(0.76767677,0.12382835), 
(0.78787879,0.11237464), 
(0.80808081,0.10106685), 
(0.82828283,0.08989955), 
(0.84848485,0.07886763), 
(0.86868687,0.06796627), 
(0.88888889,0.05719096), 
(0.90909091,0.04653741), 
(0.92929293,0.03600159), 
(0.94949495,0.02557969), 
(0.96969697,0.01526807), 
(0.98989899,0.00506332) 
}

\def\zdtpareto{
 (0.03,0.95),
 (0.15,0.75),
 (0.3,0.5),
 (0.4,0.4),
 (0.45,0.37),
 (0.57,0.3),
 (0.65,0.27),
 (0.75,0.15),
 (0.85,0.10),
 (0.98,0.02)
}

\def\zdtnoisypareto{
 (0.01,0.9),
 (0.02,0.8),
 (0.15,0.4),
 (0.23,0.37),
 (0.27,0.33),
 (0.45,0.15),
 (0.6,0.12),
 (0.65,0.10),
 (0.99,0.05)
}

\def\zdteffectivepareto{
 (0.03,0.9),
 (0.05,0.85),
 (0.3,0.6),
 (0.45,0.4),
 (0.5,0.45),
 (0.6,0.35),
 (0.7,0.25),
 (0.7,0.3),
 (0.9,0.1)
}

\begin{figure}
\centering
\scalebox{0.75}{
\begin{tikzpicture}[xscale=6,yscale=4]

\draw[-{Latex[length=30pt, width=45pt]}, blue!10!white, line width=30pt] (0.4,0.4) to node[xshift=5pt,yshift=3pt,gray,opacity=0.5,rotate=35]{\Large \textbf{Optimise}} (0.05,0.05);

\foreach \p in \zdttruepareto {
\node[circle,fill=gray,opacity=0.5,inner sep=0pt,minimum size=3pt] at \p {};
}

\foreach \p in \zdtpareto { 
\node[color=red,inner sep=0pt,opacity=0.75,mark size=3pt,line width=1pt] at \p {\pgfuseplotmark{x}};
}

\foreach \p in \zdtnoisypareto {
\node[color=blue,inner sep=0pt,opacity=0.35,mark size=1.5pt] at \p {\pgfuseplotmark{*}};
}

\foreach \p in \zdteffectivepareto {
\node[color=blue,inner sep=0pt,opacity=0.75,mark size=2pt,line width=1pt] at \p {\pgfuseplotmark{asterisk}};
}

\draw[-latex,blue,opacity=0.45] (0.01,0.9)  -- (0.03,0.9);
\draw[-latex,blue,opacity=0.45] (0.02,0.8)  -- (0.05,0.85);
\draw[-latex,blue,opacity=0.45] (0.15,0.4)  -- (0.3,0.6);
\draw[-latex,blue,opacity=0.45] (0.23,0.37) -- (0.45,0.4);
\draw[-latex,blue,opacity=0.45] (0.27,0.33) -- (0.5,0.45);
\draw[-latex,blue,opacity=0.45] (0.45,0.15) -- (0.6,0.35);
\draw[-latex,blue,opacity=0.45] (0.6,0.12)  --  (0.7,0.25);
\draw[-latex,blue,opacity=0.45] (0.65,0.10) -- (0.7,0.3);
\draw[-latex,blue,opacity=0.45] (0.99,0.05) -- (0.9,0.1);

\draw[-latex,thin,black] (-.02,0) -- (1,0) node[pos=0.5,below] {$y_0$};
\draw[-latex,thin,black] (0,-.02) -- (0,1) node[pos=0.5,left] {$y_1$};

\node[rectangle, draw=black, anchor=north east,align=left] at (1,1) {
\tiny Ideal Pareto front \graypoint \\
\tiny Pareto front \redcross \\
\tiny Noisy Pareto front \bluepoint \\
\tiny Effective Pareto front \bluestar
};

\end{tikzpicture}
}
\caption{Standard, Noisy and Effective Pareto front for ZDT1~\cite{ZDT2000}, showing the problem of noise and outliers in \ac{moo}.}
\label{fig:pareto}
\end{figure}